\documentclass[10pt,twocolumn]{IEEEtran}
\usepackage{graphicx}          
\usepackage{amsmath}
\usepackage{tabularx}
\usepackage{amsfonts}
\usepackage{cite}
\usepackage{url}
\usepackage{verbatim}
\usepackage{booktabs}
\usepackage{times,epsfig}
\usepackage{makecell}
\usepackage{graphicx}
\usepackage[normalem]{ulem}
\usepackage{multirow}
\usepackage{graphicx}
\usepackage[normalem]{ulem}
\useunder{\uline}{\ul}{}
\usepackage{psfrag}
\usepackage{subfigure}
\usepackage{stfloats}
\usepackage{setspace}
\usepackage{amssymb}
\usepackage{multirow}
\usepackage{colortbl}
\usepackage[justification=centering]{caption}
\usepackage{fancyhdr}
\usepackage[table,xcdraw]{xcolor}

\begin{document}
	
	\pagenumbering{arabic}
	
	\title{When Large Language Model Agents Meet 6G Networks: Perception, Grounding, and Alignment}
	\author{Minrui Xu, Dusit Niyato, \emph{Fellow, IEEE}, Jiawen Kang, Zehui Xiong, Shiwen Mao, \emph{Fellow, IEEE}, \\Zhu Han, \emph{Fellow, IEEE}, Dong In Kim, \emph{Fellow, IEEE}, and Khaled~B.~Letaief, \emph{Fellow, IEEE}
 \thanks{Minrui~Xu and Dusit~Niyato are with the School of Computer Science and Engineering, Nanyang Technological University, Singapore 639798, Singapore (e-mails: minrui001@e.ntu.edu.sg; dniyato@ntu.edu.sg). Jiawen~Kang is with the School of Automation, Guangdong University of Technology, China (e-mail: kavinkang@gdut.edu.cn). Zehui~Xiong is with the Pillar of Information Systems Technology and Design, Singapore University of Technology and Design, Singapore 487372, Singapore (e-mail: zehui\_xiong@sutd.edu.sg). Shiwen~Mao is with the Department of Electrical and Computer Engineering, Auburn University, Auburn, AL 36849-5201 USA (smao@ieee.org). Zhu~Han is with the Department of Electrical and Computer Engineering, University of Houston, Houston, TX 77004 USA, and also with the Department of Computer Science and Engineering, Kyung Hee University, Seoul 446-701, South Korea (zhuhan22@gmail.com). Dong In Kim is with the Department of Electrical and Computer Engineering, Sungkyunkwan University, Suwon 16419, South Korea  (e-mail: dikim@skku.ac.kr). Khaled B. Letaief is with the Department of Electrical and Computer Engineering, Hong Kong University of Science and Technology (HKUST), Hong Kong (e-mail: eekhaled@ust.hk).}
 }
	\maketitle
	\pagestyle{headings}

	\begin{abstract}
AI agents based on multimodal large language models (LLMs) are expected to revolutionize human-computer interaction and offer more personalized assistant services across various domains like healthcare, education, manufacturing, and entertainment. Deploying LLM agents in 6G networks enables users to access previously expensive AI assistant services via mobile devices democratically, thereby reducing interaction latency and better preserving user privacy. Nevertheless, the limited capacity of mobile devices constrains the effectiveness of deploying and executing local LLMs, which necessitates offloading complex tasks to global LLMs running on edge servers during long-horizon interactions.
In this article, we propose a split learning system for LLM agents in 6G networks leveraging the collaboration between mobile devices and edge servers, where multiple LLMs with different roles are distributed across mobile devices and edge servers to perform user-agent interactive tasks collaboratively. In the proposed system, LLM agents are split into perception, grounding, and alignment modules, facilitating inter-module communications to meet extended user requirements on 6G network functions, including integrated sensing and communication, digital twins, and task-oriented communications. Furthermore, we introduce a novel model caching algorithm for LLMs within the proposed system to improve model utilization in context, thus reducing network costs of the collaborative mobile and edge LLM agents.
	\end{abstract}

	\begin{IEEEkeywords}
6G networks, AI agents, integrated sensing and communication, digital twins, task-oriented communications
	\end{IEEEkeywords}
	
\section{Introduction}

AI agents, designed to integrate AI models into everyday services as personal assistants to humans, have become a pivotal element in advancing towards artificial general intelligence (AGI) \cite{wang2023survey, xi2023rise}. AI agents powered by large language models (LLMs), i.e., LLM agents, possess the capability to follow user instructions, observe environments, make decisions, and execute actions at a human-equivalent level. Therefore, LLM agents can proactively provide users with recommendations for final decisions by understanding and remembering cross-application user intentions and behaviors.
Particularly, AI agents observe surrounding environments by processing information in various modalities from sensors, leveraging the versatility of multimodal LLMs~\cite{yang2023dawn}. In addition, LLM agents can solve complex tasks by grounding the plan of action to achieve their missions through reasoning, memory, and verification. After the alignment between LLM agents and humans, agents can attain human-like intelligence to provide recommendations to users with text, tools, and embodied actions that are consistent with human values.

Although the deployment of LLM agents on mobile devices in 6G networks allows democratizing access to services currently considered prohibitively expensive at cloud data centers, several issues remain in implementing the LLM agents for complex, multi-round interaction agent services~\cite{shen2024large, lin2023pushing}. For mobile devices with limited capacities, running AI models of agents, which is both computation- and memory-intensive, is challenging for supporting the long-term execution of LLMs. In addition, these limitations are further exacerbated by the restricted context windows of LLMs, hindering LLM agents from performing long-term and complex interactions, such as perception, reasoning, and coding, which consume considerable available context resources~\cite{wu2023autogen}. To address these challenges, a split learning system based on collaborative end-edge-cloud computing, which aims at partitioning LLM agents into mobile and edge agents, emerges as a viable solution. In this system, mobile LLM agents, operating local LLMs (0-10B parameters, e.g., LLAMA-7B) on mobile devices, can handle real-time, direct perception and alignment tasks. Meanwhile, edge LLM agents, hosting global LLMs, ($>$10B parameters, e.g., GPT-3) on edge servers, can utilize global information and historical memory to help mobile LLM agents perform complex tasks.

There are several advantages to partitioning LLM agents into mobile and edge agents in 6G networks. First, flexible deployment of LLM agents can be supported by heterogeneous devices with different locations, capabilities, and contextual adaptability. Specifically, mobile LLM agents with proper local LLMs can operate effectively with their computing capabilities regardless of their locations and user scenarios. Second, long-horizon collaboration can be enabled across multiple mobile devices by bridging the integration between low-level operational plans of local LLMs and high-level strategic plans of global LLMs. Third, mobile LLM agents exhibit enhanced adaptability in dynamic open-ended environments. For instance, mobile LLM agents can understand instructions using local LLMs and then adjust their actions based on immediate environmental feedback for real-time responsiveness and relevance during their interactions with physical environments.

In this article, we propose a split learning system of LLM agents consisting of mobile LLM agents and edge LLM agents in 6G networks, which is democratic, flexible, and long-horizon for running sustainable AI agents in open-ended environments. First, we introduce the basic concept of AI agents and introduce the processes of constructing LLM agents via collaborative end-edge-cloud computing. Secondly, we discuss three main issues in developing LLM agents in 6G networks, including multimodal perception, interactive grounding, and alignment with humans. Thirdly, we investigate a real-world application that leverages mobile and edge LLM agents to generate accident reports collaboratively. At an accident site, vehicles can employ mobile LLM agents to observe the surrounding scene of a car accident and generate their local environmental descriptions. By sending these descriptions to edge servers, edge LLM agents can use global observations to deduce and offer more detailed and precise plans for vehicles. Finally, the mobile LLM agents can generate text responses, functional call requests, and embodied actions based on the global plan. In addition, we propose a metric called \textit{age of thought} (AoT) to assess the significance of thoughts, i.e., the intermediate steps generated by LLMs, during the reasoning and planning processes of edge LLM agents. This metric emphasizes that older thoughts hold less importance and thus can ensure the high performance of cached models. Based on this metric, we introduce the Least Age-of-Thought (LAoT) model caching algorithm, which evicts global models that have the least impactful and relevant thoughts, and thus reduces the grounding cost in terms of latency, resource consumption, and performance loss for serving edge LLM agents in 6G networks.
Overall, our main contribution can be summarized as follows.
\begin{itemize}
    \item We propose a split learning system for LLM agents in 6G networks, which aims to provide democratic AI assistant services via the collaboration of mobile and edge LLM agents over end-edge-cloud computing.
    \item During the integration of 6G networks and LLM agents, we discuss several major issues, including integrated sensing and communication for multimodal perception, digital twins for grounding decisions, and task-oriented communications for the alignment of agents.
    \item We propose a new optimization framework in the system, i.e., model caching for AI agents, which aims at maximizing the in-context learning capabilities of LLM agents while reducing the network costs of serving mobile and edge LLM agents.
\end{itemize}









\section{Collaborative end-edge-cloud computing for LLM agents in 6G Networks}
As a pivotal stride towards achieving AGI, AI agents are the key computational entities that can proactively perceive user instructions, observe the environment, ground decisions, and perform human-like actions~\cite{xi2023rise}. In 6G networks, AI agents are developed to execute intricate tasks collaboratively, from managing networks to acting as personal assistants for humans. According to the difference in fundamental working mechanisms, there are two major categories of AI agents, i.e., reinforcement learning (RL) agents and LLM agents, which are discussed below.

\subsection{Categories of AI Agents}

\subsubsection{RL Agents} 

Utilizing RL algorithms to observe states, make decisions, and take actions in an environment, RL agents learn through trial and error, by receiving feedback as rewards or penalties as a result of their actions. They aim to maximize their cumulative reward over time by learning optimal policies. For example, in communications and networking, RL agents can make decisions for dynamic network access, transmit power control, wireless caching, and data offloading locally to maximize network performance under uncertain network environments. 
Specifically, RL agents formulate the communication and networking environment into a Markov decision process (MDP) consisting of states, actions, transition probabilities, and rewards. 
However, although RL agents learn to make decisions for network access and management~\cite{xi2023rise}, they cannot interact with humans and other agents using texts in open-ended environments which limits their potential to offer more diverse services that require understanding and responding to human instructions.

\subsubsection{LLM Agents}

To achieve human-level intelligence, LLM agents build upon versatile and powerful LLMs that have demonstrated remarkable capabilities in few-shot and zero-shot environment perception and instruction understanding~\cite{wang2023survey, xi2023rise}. In addition to the decision-making capabilities of RL agents, LLM agents can interact with the environment through texts, API tools, and embodied actions continuously while gradually improving their performance during the interaction. Meanwhile, pre-training on large-scale datasets elicits emerging abilities of LLMs, allowing them to tackle various downstream tasks related to data management, question answering, route planning, and scientific inquiries. Furthermore, equipped with memory, reasoning, planning, and tool capabilities, LLM agents can not only make decisions for network environments but also leverage language understanding and employ tools such as the Internet and databases for tackling complex control tasks. Compared with the generalization of RL agents, the role-playing capability of LLM agents allows them to serve specific roles while handling different tasks. For example, LLM agents can act as experiment assistants, automating the design, planning, and execution of scientific experiments based on human-crafted instructions. However, textual instructions are usually not sufficient for LLM agents to perceive the entire environment in a realistic setting.

To enhance LLMs with multi-sensory capabilities, such as visual and audio understanding, multimodal LLMs~\cite{yang2023dawn}, like GPT-4V(ision), are introduced for agents to perceive and process inputs from multiple modalities, including tactile feedback, gestures, Inertial Measurement Units (IMUs) motion sensor data, and 3D maps. For visual input, multimodal LLMs can be leveraged to generate a description for the current environment, where they can produce multimodal descriptions, such as text, audio, and images, which enable better accessibility for visually impaired individuals and improve positioning capabilities. Specifically, multimodal LLM agents can use a pre-trained encoder to convert signals from different modalities into a common textual representation, allowing for reasoning across modalities~\cite{moon2023anymal}. 

\subsection{Construction of Mobile Edge-empowered Agents}

\begin{figure*}
    \centering
    \includegraphics[width=0.9\linewidth]{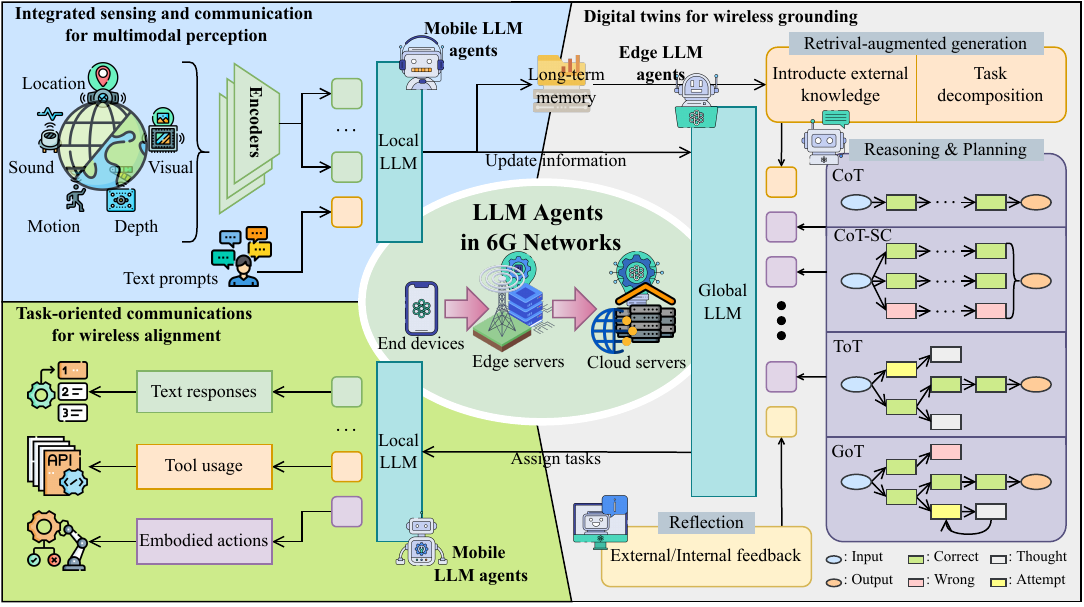}
    \caption{The split learning system of mobile and edge LLM agents over collaborative end-edge-cloud computing.}
    \label{fig:agent}
\end{figure*}

As illustrated in Fig.~\ref{fig:agent}, the construction of LLM agents in collaborative end-edge-cloud computing consists of three main processes, namely, mobile LLM agent execution, edge LLM agent execution, and inter-agent communication between mobile agents and edge agents to update information and assign tasks.

\subsubsection{Mobile LLM Agent Execution}

Initially, each user downloads tiny local LLMs (0-10B), e.g., LLAMA-7B, to its mobile device from edge servers via radio access networks (RANs) for personalized initialization. During initialization, users can configure mobile LLM agents with personal profiles such as age, gender, and career, which agents use to tailor their interactions and responses with specific roles. In addition, mobile LLM agents can leverage contextual initialization based on the current situation by processing and analyzing historical interactions. There are two major methods for LLM agents to perceive environments, i.e., human instruction and sensing. On the one hand, human instructions are given through interactive dialogues between humans and LLM agents. On the other hand, LLM agents can perceive the physical environment, which provides multimodal sensory inputs from interacting objects, including visual, auditory, and spatial data.

To process the received instruction and multimodal sensing data, mobile LLM agents can utilize pre-trained components, such as modality encoders, word embedding layers, and projection layers, to combine multi-sensory inputs. Each modality encoder is specific to one modality, such as CLIP for images, CLAP for audio signals, IMU2CLIP for IMU motion sensor, and Intervideo for videos~\cite{girdhar2023imagebind}. In mobile devices, multiple encoders process and combine the multimodal input data and then project the output into the text token embedding space of local LLMs. To process human instructions, the word embedding layer is a crucial component that maps words or tokens into a continuous vector space, capturing semantic relationships between them, and helping in understanding user-specific instructions. In mobile AI agents, due to limited capacities in mobile devices, tiny local LLMs with a limited amount of parameters can generate real-time responses based on local perception but cannot tackle complex tasks that require comprehensive consideration and generalization. 

\subsubsection{Global Agent Execution}

In edge servers, edge LLM agents with huge global LLMs ($>$10B), e.g., GPT3, can leverage long-term memory, reasoning, and planning modules to enhance the quality of responses with global information and understanding of environments. Historical interactions of mobile LLM agents can be stored as long-term memory in vector databases through memory embedding layers. Based on the long-term memory from mobile LLM agents, edge LLM agents can use retrieval-augmented generation (RAG) to output responses with better performance and consistency~\cite{asai2023retrieval}. In addition, edge LLM agents can utilize chain-of-thought (CoT) reasoning to improve the performance in complex tasks~\cite{chu2023survey}. When tackling complex tasks, edge LLM agents using CoT start by employing various reasoning paths to deduce potential answers, considering that each complex problem has multiple ways of thinking. This way, edge LLM agents can adapt to unfamiliar scenarios through knowledge generalization and transfer abilities inherent in global LLMs. Furthermore, edge LLM agents can leverage self-reflection to verify reasoning paths, gaining more accurate results of their actions and making better decisions for future behaviors.

\subsection{Inter-agent Communication between Local and Edge Agents}

When mobile LLM agents are incapable of accomplishing complex tasks, they can offload the intermediate results, including local perceptions and user intentions, to edge LLM agents equipped with huge global LLMs and global information for remote execution. Mobile LLM agents can transmit intermediate results, such as text or other embeddings, through inter-agent communication over RANs. Due to limited bandwidth and uncertain wireless channels, mobile LLM agents need to optimize the size of the transmitted content, i.e., intermediate inference results of local LLMs, and configure communication parameters for successful offloading, e.g., the transmit power and the chosen channel. Based on the responses and decision results generated by edge LLM agents, mobile LLM agents adapt global general plans to the local specific plans to interact with users and the environments. After understanding the locally specific plans using local tiny LLMs, mobile LLM agents generate responses, use API tools, and perform embodied actions locally using their actuation modules.




\section{ISAC for Wireless Perception: Ubiquitous and Adaptability}

To run LLM agents efficiently in 6G networks with ubiquitous low-end devices, mobile LLM agents can perceive user instructions and sense environments for modeling and understanding the current situation. In addition, to improve adaptability and generalization, mobile LLM agents need to offload computation-intensive and intractable tasks to edge LLM agents for remote execution. Therefore, mobile LLM agents need to collect and extract information from noisy observations and communicate with edge servers to transfer information, which requires the implementation of integrated sensing and communication (ISAC) by utilizing the wireless {\color{blue}communication} infrastructure.

\subsubsection{Environmental Perception} 

In multi-functional 6G networks, mobile LLM agents can autonomously perceive the surrounding environment using equipped sensors~\cite{moon2023anymal}, which consume network resources for supporting the sensing functionality. By integrating basic perceptual abilities such as vision, text, and light sensitivity, LLM agents can develop various user-friendly perception modules~\cite{liu2022integrated}. For example, LLM agents in mobile devices can perceive more complex user inputs, such as eye-tracking, body motion capture, and even brainwave signals in brain-computer interaction. Furthermore, LLM agents in vehicular networks can be equipped with Lidar, GPS, and IMUs, allowing them to perceive location-based data for vehicles and mobile users.

\subsubsection{Human-language Instruction}

During the interaction between users and agents, text instructions can be given to mobile LLM agents by providing them with explicit requests as well as implied values and intentions. Mobile LLM agents can understand implicit meanings within textual input based on contextual interaction with users, thanks to their short-term memory. After processing through local LLMs, mobile LLM agents can respond with answers in human language. Additionally, users can also provide instructions via audio, which contains environmental information compared to text~\cite{moon2023anymal}. Handling audio input involves leveraging existing models, cascading paradigms, and integrating audio with other modalities to enhance agents' perception and understanding of the environment.

\subsubsection{Inter-agent Interactions}

In the proposed system, ubiquitous interaction between mobile and edge LLM agents is crucial for offloading intermediate results, receiving feedback, interactive reasoning, and self-reflection over RANs~\cite{wu2023autogen}. During collaboration between mobile and edge agents, they need to continuously communicate with each other with messages in text or other embedded formats in a noisy environment. Therefore, this communication process usually consumes a large amount of bandwidth and network resources for long-term and multimodal interactions.

For the wireless perception of LLM agents in multi-functional 6G networks, ISAC is promising to improve spectral and energy efficiencies for mobile LLM agents to collect information from environments and transmit intermediate results to edge LLM agents simultaneously. For example, mobile LLM agents in vehicles need to perform radar sensing and transmit the perception results to edge LLM agents simultaneously. By utilizing network resources more efficiently for sensing and communication, ubiquitous LLM agents can be deployed in wireless environments and become more adaptable to a dynamic and open-ended world.

\section{Digital Twins for Wireless Grounding: Reliability and Consistency}

For grounding the responses and actions, mobile LLM agents maintain digital twins (DTs) at the edge servers to interactively perform retrieval-augmented generation (RAG), reasoning and planning, and reflection with edge LLM agents in 6G networks with hyper reliable and low-latency communication. DTs of mobile LLM agents are created as digital replicas of physical entities with perceived data and can help to perform global grounding with internal memory and external knowledge. Through continuously updating external observation and internal reasoning results, DTs of mobile LLM agents can be created for real-time monitoring, analysis, and optimization of the decisions of mobile agents while performing complex tasks.

\subsubsection{Memory and Retrieval-Augmented Generation}

In the proposed system, mobile LLM agents maintain a short-term memory while global agents maintain a long-term memory for grounding agents' responses and actions. In mobile LLM agents, the short-term memory is collected through various mechanisms, such as in-context learning, maintaining internal states, utilizing scene descriptions or environment feedback, and generating task plans. Additionally, short-term memory in mobile LLM agents can be converted to long-term memory by leveraging external storage resources in edge servers, such as vector databases, that allow rapid querying and retrieval of information as needed. Based on the long-term memory, edge LLM agents can perform RAG~\cite{asai2023retrieval} to improve consistency in generation by using a retrieval model to retrieve relevant information from a knowledge base or a set of reference documents and then incorporating this retrieved information into the generation process. In this way, by processing retrieved content using global LLMs, RAG can be leveraged in complex and long-horizon tasks using specialized knowledge, up-to-date information, and customizable definitions for better performance and consistency. Specifically, edge LLM agents can access past responses of mobile LLM agents to ensure consistent collaboration.

\subsubsection{Reasoning and Planning}

Edge LLM agents can tackle complex tasks by decomposing them into sequential steps and sub-tasks to output accurate responses. To perform intricate reasoning, CoT~\cite{chu2023survey} involving a step-by-step reasoning process along a single path can improve the reliability and interpretability of LLMs decisions. Specifically, edge LLM agents have the ability to use CoT to break down complex tasks and offer step-by-step instructions for mobile LLM agents to complete individual tasks. In addition, self-consistent CoT (CoT-SC) is proposed to improve performance of reasoning tasks by aggregating multiple language model outputs and selecting the most consistent answer through a majority vote. To extend CoT, tree-of-thoughts (ToT), a proposed extension of CoT that formulates thought units into a tree structure, allows LLMs to explore coherent thought units as intermediate steps, enabling better problem-solving and planning capabilities. Moreover, graph-of-thoughts (GoT) is a static structure that specifies the graph decomposition of a given task in the CoT paradigm. It prescribes the transformations to be applied to language model thoughts, along with their order and dependencies. Although these step-by-step reasoning and planning mechanisms allow for multiple choices at each step and mimic human thinking, they might request more computing resources from edge servers to generate the intermediate results compared with outputting the results directly. 

\subsubsection{Verification and Reflection}

To ensure the correctness of the reasoning process before final response generation, LLM agents can leverage verification reflection to validate the correctness of each step in the CoT process. For example, SelfCheck~\cite{xi2023rise} is a zero-shot checking scheme for LLMs that aims to improve question-answering accuracy by identifying errors in the LLM's reasoning process. It works as a step-by-step checker, individually checking each step in the LLM's reasoning process based on the available context. Using confidence scores as weights, SelfCheck allows for improved question-answering accuracy by focusing on the most accurate answer. Therefore, LLM agents can independently summarize and infer more abstract, complex, and high-level information.

Therefore, there should be a tradeoff between the performance of grounding modules and the computing resources consumed during the grounding processes. For inter-agent grounding, mobile and edge LLM agents can leverage their own self-correction capabilities to improve the accuracy of final decisions leveraging computing resources in mobile devices and edge servers. Moreover, inter-agent grounding requires additional networking resources to transmit correction results between mobile and edge LLM agents for cross-verification for grounding the actions of LLM agents.


\section{Task-oriented Communications for Wireless Alignment: Trustworthy and Generalizability}

In 6G networks with limited bandwidth resources, task-oriented communications~\cite{10183789} refer to a communication approach where the performance is measured based on the success level of achieving a sequence of application-related tasks, rather than traditional metrics such as data rate or wireless link reliability. The alignment of LLM agents in 6G networks offering ubiquitous connectivity can be regarded as a type of task-oriented communication where LLM agents can leverage the massive resources of mobile devices and edge servers to achieve their alignment goals. For instance, mobile LLM agents are the data destinations of global general plans from edge LLM agents to generate texts, call API functions, and perform embodied actions. In addition, feedback from humans and mobile LLM agents can be collected as datasets for supervised fine-tuning, reinforcement learning from human feedback (RLHF), and direct preference optimization (DPO) to regularize global LLMs. Beyond data-oriented communications, the accomplishment of alignment between mobile and edge LLM agents and humans in view of semantics is directly linked to task-oriented communication performance and strategies that wireless users and mobile and edge LLM agents can provide real-time evaluation and feedback. 

\subsubsection{Text Responses}

Since LLMs are pre-trained on large-scale datasets with biased data, a mismatch or distribution shift between the training and test data can cause LLMs to generate incorrect information, known as hallucination~\cite{liu2023trustworthy}. However, to ensure that mobile and edge LLM agents align with human intentions and preferences, the system needs to reduce the likelihood of generating harmful outputs and improve usability by better following human instructions. For instance, OpenAI (https://openai.com/blog/introducing-superalignment), the creator of ChatGPT, has announced that they are going to leverage 20\% of computing resources to fine-tune strong pretrained LLMs for regularizing LLMs to faithfully follow instructions or generate safe outputs. Fortunately, wireless alignment enables massive users to contribute their efforts and computing resources in alignment activities and contribute their efforts towards unlocking the full potential of LLMs while following human value and intentions, positively impacting various domains and enriching human experiences.

\subsubsection{Tool Usage and Generation}

By instruction fine-tuning on API datasets, mobile LLM agents should become proficient in leveraging tools and APIs to accomplish intricate tasks and interact with different virtual applications effectively based on general plans from edge LLM agents~\cite{qin2023toolllm}. Therefore, depending on specific environments and agent types, mobile LLM agents can be customized using local instruction fine-tuning datasets, which encompass real-world APIs and practical scenarios, to accomplish both single-tool and multi-tool tasks. Furthermore, mobile LLM agents can showcase exceptional adaptability when faced with unfamiliar APIs and tool-use datasets that are outside their usual field, particularly when users are inexperienced in the given environments.

\subsubsection{Embodied Actions}

To effectively interact with humans and physical environments, mobile LLM agents can perform embodied actions, including movements, gestures, or other physical behaviors~\cite{xi2023rise} to interact with the physical world directly under the high-level plans from edge LLM agents. Embodied actions of mobile LLM agents are physically performed according to their design, mechanical features, and technology. For instance, vehicles can perform basic mechanical movements, such as driving forward and backward, tuning, braking, and accelerating. By performing these embodied actions, vehicles can easily adapt to road conditions, e.g., bumpy roads, slippery surfaces, and bad weather by adjusting internal temperature and air quality. To navigate through unfamiliar environments, robots with mobile LLM agents need to gather information, carry out tasks, and interact with other agents like humans. By performing embodied actions, mobile and edge LLM agents can extend their capabilities beyond digital boundaries, and interact and manipulate their physical surroundings directly.
\begin{figure*}
    \centering
    \includegraphics[width=0.9\linewidth]{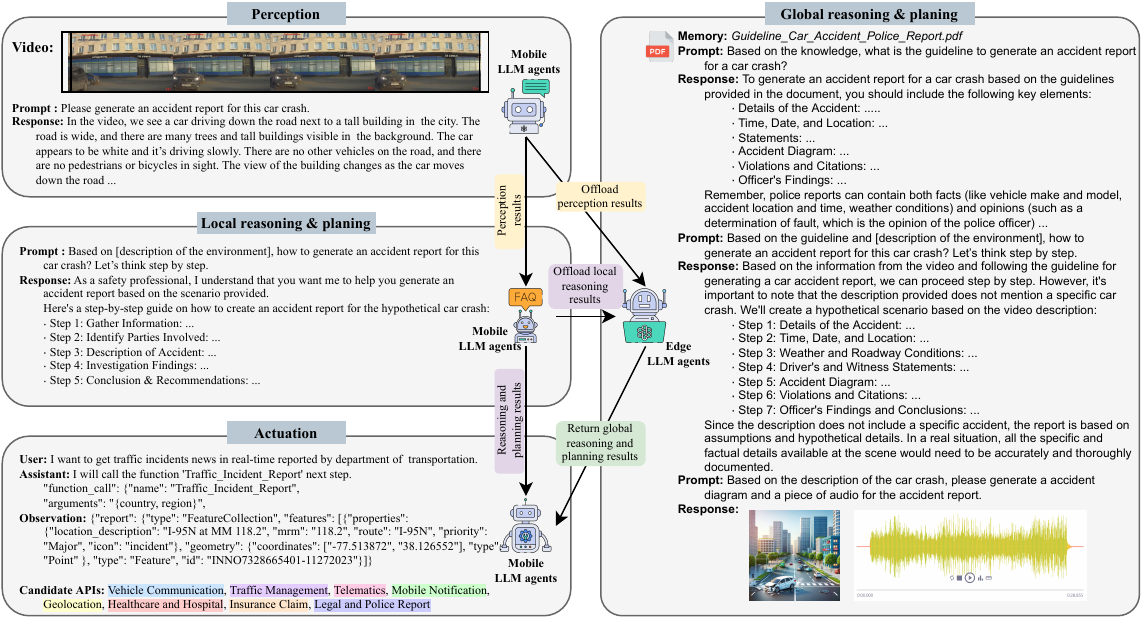}
    \caption{An example of mobile and edge LLM agents used in the generation of accident reports for car crashes.}
    \label{fig:example}
\end{figure*}
\section{Case Study of Model Caching for Collaborative Mobile and Edge LLM agents}

In the split learning system of LLM agents over collaborative end-edge-cloud computing, each mobile AI agent is composed of a perception module, a local reasoning module, and an alignment module while each edge LLM agent consists of a global reasoning and planning module. In addition to allocating traditional computing, communication, and storage resources for executing LLM agents, the LLMs running in these agents are new resources to be allocated for performing contextual tasks of AI agents. Specifically, mobile LLM agents can leverage local LLMs for zero-shot environmental perception and auction, which is more comprehensive. Meanwhile, edge LLM agents with global LLMs can perform more intricate step-by-step reasoning and planning with global information for reliable and interpretative decision-making.

To construct the perception module of mobile LLM agents that can collect multimodal information from the environment, we leverage the ImageBind~\cite{girdhar2023imagebind} and the LanguageBind~\cite{zhu2023languagebind} for sensing the environment. The ImageBind, based on ViT-Huge, and the LanguageBind, based on ViT-Large, use image embeddings as a central anchor to align embeddings from other modalities like text, audio, depth, thermal, and IMU data. The LanguageBind employs contrastive learning to align and bind different modalities including video, infrared, depth, and audio from the environment with the language modality. In this study, we evaluate the perception module using the IN1K dataset for image data, the K400 dataset for video data, the LLVIP dataset for infrared data, the NYU-D dataset for depth data, the ESC-50 dataset for audio data, and the Ego4D dataset for IMU data.


In this use case, we leverage mobile and edge LLM agents to generate accident reports collaboratively for car crashes, where multiple mobile agents perceive the environment and report their local observations to edge LLM agents. Meanwhile, edge LLM agents generate a general report by leveraging global information and high-level plans for mobile agents.
We validate the performance of LLM agents based on the Car Crash dataset (https://github.com/Cogito2012/CarCrashDataset). The mobile LLM agents on vehicles are implemented based on LLaMA. The perception module is developed based on Video-LLaMA, the local grounding module is developed based on LLaMA-7B-Chat, and the alignment module is developed based on ToolLLM. The edge LLM agent on the edge server is developed based on GPT4 and we implement a GPT named ``Accident Report Assistant" as the global AI agent (https://chat.openai.com/g/g-7sWJT5dSD-accident-report-assistant). The actuation module is implemented with ToolLLaMA, which is a fine-tuned LLaMA-7B model using the instruction-solution pairs.

\begin{table}[t]
\small\centering
\caption{Evaluation of mobile LLM agents with different perception modules for different modalities.}
\label{tab:perception}
\begin{tabular}{|c|c|c|}
\hline
         & ImageBind        & LanguageBind        \\ \hline
Model            & ViT-Huge (632 M) & ViT-Large/14 (307M) \\ \hline
Image - IN1K     & 77.7             & -                   \\ \hline
Video - K400     & 50.0             & 64.0                \\ \hline
Infrared - LLVIP & 63.4             & 87.2                \\ \hline
Depth - NYU-D    & 54.0             & 65.1                \\ \hline
Audio - ESC-50   & 66.9             & 89.8                \\ \hline
IMU - Ego4D      & 25.0             & -                   \\ \hline
\end{tabular}
\end{table}

After the perception of environments, each mobile LLM agent for perception describes the local situation and reports the current situation to the edge LLM agent. Based on the collected local perception, the edge LLM agent aggregates them into a comprehensive picture for the following multi-step reasoning and planning processing, e.g., CoT. Finally, the edge LLM agent provides general plans to mobile LLM agents for actuation and lets them interact with users and environments with local plans translated by their local LLMs, including text responses, APIs, and embodied actions. Due to the limited context window of mobile and edge LLM agents, we consider the inference of perception and actuation to be zero-shot and their performance is determined by perception fidelity of multimodal information and successful ratio during interaction with users and environments. Furthermore, edge LLM agents can provide suggestions for multiple local agents and their performance is affected by their historical thoughts. As the CoT is a step-by-step inference process that generates multiple intermediate thoughts during the grounding of final decisions, the thought that is closer to the final decisions should contribute more value to making the final decisions. In this regard, we propose a metric of age-of-thought (AoT) to evaluate the value of thoughts based on their freshness. 

With the limited memory of edge servers and the massive amount of parameters of LLMs, edge servers cannot load all the models into the main memory at the same time. To provide AI services to satisfy user requirements, edge servers need to schedule the global AI models for reasoning and planning for the requested services. To minimize the cost in terms of edge accuracy loss, model switching cost, edge inference cost, edge inference latency, and cloud inference cost, effective model caching algorithms should be designed to manage loaded models for edge LLM agents. Especially, the cached models not only can be evicted proactively according to the caching policies, but also can be evicted due to the used context exceeding the context window. Therefore, we propose an LAoT model caching algorithm based on the concept of AoT where the model with the least valuable thoughts is evicted.

\begin{figure}
\vspace{-0.3cm}
    \centering
    \includegraphics[width=0.8\linewidth]{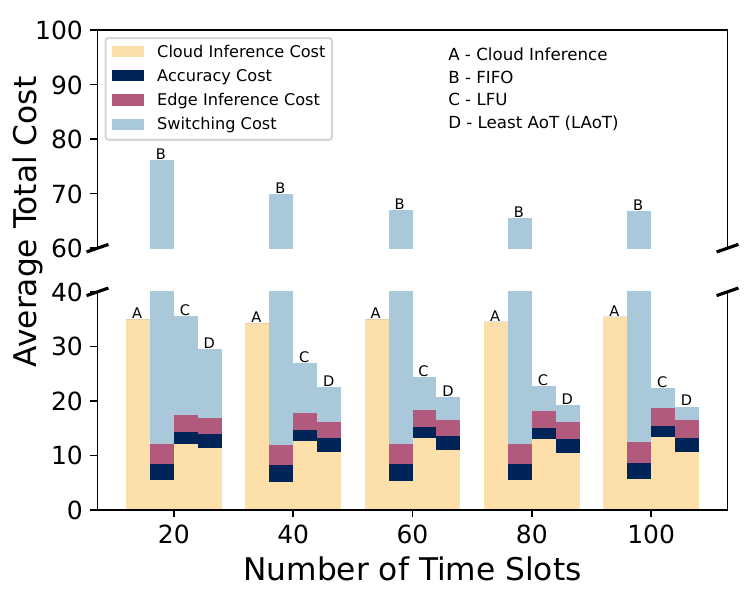}
    \caption{The execution cost of various model caching algorithms in different time slots.}
    \label{fig:exp}
\end{figure}

The maximum token consumption for each CoT step is set to 200 in the experiment. The context window of LLaMA is 4K tokens, the context window of GPT-3.5-turbo is 16K tokens, and the context window of GPT-4 is 32K tokens. We consider an edge server with 64 GPUs with 80 GB memory, 312 TFLOPS, and 300W max thermal design power. We consider 30 types of services and 10 edge LLM agents. The experimental results are illustrated in Fig.~\ref{fig:exp}. As the number of time slots increases, the cost of edge inference of LLM agents decreases due to less switching cost and higher model performance. The reason is that the popular models are loaded into the memory of edge servers. In addition, during the inference of edge LLM agents, the thoughts are accumulated to improve the reasoning and planning results and thus the edge accuracy loss is lower. Overall, we can observe that the LC algorithm can improve the accuracy of edge LLM agents while reducing total execution costs compared to the existing baselines, including cloud-only inference, the first-in-first-out (FIFO) policy, and the least frequently used (LFU) policy.





\section{Conclusions and Future Directions}
In this article, we have proposed a split learning system for LLM agents over collaborative end-edge-cloud computing in 6G networks for multimodal perception, interactive grounding, and alignment. We have introduced the evolution of LLM agents and the construction of LLM agents over end-edge-cloud computing with collaborative mobile and edge LLM agents. Furthermore, we have investigated the communication and networking issues in developing mobile edge-empowered agents including perception, grounding, and alignment. Finally, we have developed a use case for the application of mobile and edge LLM agents in vehicular networks and propose a model caching algorithm to optimize the performance of AI agent services while reducing execution costs.

In future research, it is important to explore further integration of 6G networks and AI agents. This could involve incorporating next-generation multiple access, metasurface, and over-the-air computation to support LLM agents in dynamic wireless environments. Additionally, it is crucial to address the model privacy concerns that may arise during collaboration between mobile and edge LLM agents. This will help prevent any potential information breach, especially in cases where malicious edge servers may attempt to access users' private information from running models.

\bibliographystyle{ieeetr}
\bibliography{main}

\begin{thebibliography}{10}

\bibitem{wang2023survey}
L.~Wang, C.~Ma, X.~Feng, Z.~Zhang, H.~Yang, J.~Zhang, Z.~Chen, J.~Tang, X.~Chen, Y.~Lin, {\em et~al.}, ``A survey on large language model based autonomous agents,'' {\em arXiv preprint arXiv:2308.11432}, 2023.

\bibitem{xi2023rise}
Z.~Xi, W.~Chen, X.~Guo, W.~He, Y.~Ding, B.~Hong, M.~Zhang, J.~Wang, S.~Jin, E.~Zhou, {\em et~al.}, ``The rise and potential of large language model based agents: A survey,'' {\em arXiv preprint arXiv:2309.07864}, 2023.

\bibitem{yang2023dawn}
Z.~Yang, L.~Li, K.~Lin, J.~Wang, C.-C. Lin, Z.~Liu, and L.~Wang, ``The dawn of lmms: Preliminary explorations with gpt-4v (ision),'' {\em arXiv preprint arXiv:2309.17421}, 2023.

\bibitem{shen2024large}
Y.~Shen, J.~Shao, X.~Zhang, Z.~Lin, H.~Pan, D.~Li, J.~Zhang, and K.~B. Letaief, ``Large language models empowered autonomous edge ai for connected intelligence,'' {\em IEEE Communications Magazine}, 2024.

\bibitem{lin2023pushing}
Z.~Lin, G.~Qu, Q.~Chen, X.~Chen, Z.~Chen, and K.~Huang, ``Pushing large language models to the 6{G} edge: Vision, challenges, and opportunities,'' {\em arXiv preprint arXiv:2309.16739}, 2023.

\bibitem{wu2023autogen}
Q.~Wu, G.~Bansal, J.~Zhang, Y.~Wu, S.~Zhang, E.~Zhu, B.~Li, L.~Jiang, X.~Zhang, and C.~Wang, ``Autogen: Enabling next-gen llm applications via multi-agent conversation framework,'' {\em arXiv preprint arXiv:2308.08155}, 2023.

\bibitem{moon2023anymal}
S.~Moon, A.~Madotto, Z.~Lin, T.~Nagarajan, M.~Smith, S.~Jain, C.-F. Yeh, P.~Murugesan, P.~Heidari, Y.~Liu, {\em et~al.}, ``Anymal: An efficient and scalable any-modality augmented language model,'' {\em arXiv preprint arXiv:2309.16058}, 2023.

\bibitem{girdhar2023imagebind}
R.~Girdhar, A.~El-Nouby, Z.~Liu, M.~Singh, K.~V. Alwala, A.~Joulin, and I.~Misra, ``Imagebind: One embedding space to bind them all,'' in {\em Proceedings of the IEEE/CVF Conference on Computer Vision and Pattern Recognition}, pp.~15180--15190, 2023.

\bibitem{asai2023retrieval}
A.~Asai, S.~Min, Z.~Zhong, and D.~Chen, ``Retrieval-based language models and applications,'' in {\em Proceedings of the 61st Annual Meeting of the Association for Computational Linguistics (Volume 6: Tutorial Abstracts)}, pp.~41--46, 2023.

\bibitem{chu2023survey}
Z.~Chu, J.~Chen, Q.~Chen, W.~Yu, T.~He, H.~Wang, W.~Peng, M.~Liu, B.~Qin, and T.~Liu, ``A survey of chain of thought reasoning: Advances, frontiers and future,'' {\em arXiv preprint arXiv:2309.15402}, 2023.

\bibitem{liu2022integrated}
F.~Liu, Y.~Cui, C.~Masouros, J.~Xu, T.~X. Han, Y.~C. Eldar, and S.~Buzzi, ``Integrated sensing and communications: Toward dual-functional wireless networks for 6{G} and beyond,'' {\em IEEE Journal on Selected Areas in Communications}, vol.~40, no.~6, pp.~1728--1767, 2022.

\bibitem{10183789}
Y.~Shi, Y.~Zhou, D.~Wen, Y.~Wu, C.~Jiang, and K.~B. Letaief, ``Task-oriented communications for 6{G}: Vision, principles, and technologies,'' {\em IEEE Wireless Communications}, vol.~30, no.~3, pp.~78--85, 2023.

\bibitem{liu2023trustworthy}
Y.~Liu, Y.~Yao, J.-F. Ton, X.~Zhang, R.~G.~H. Cheng, Y.~Klochkov, M.~F. Taufiq, and H.~Li, ``Trustworthy llms: a survey and guideline for evaluating large language models' alignment,'' {\em arXiv preprint arXiv:2308.05374}, 2023.

\bibitem{qin2023toolllm}
Y.~Qin, S.~Liang, Y.~Ye, K.~Zhu, L.~Yan, Y.~Lu, Y.~Lin, X.~Cong, X.~Tang, B.~Qian, {\em et~al.}, ``Toolllm: Facilitating large language models to master 16000+ real-world apis,'' {\em arXiv preprint arXiv:2307.16789}, 2023.

\bibitem{zhu2023languagebind}
B.~Zhu, B.~Lin, M.~Ning, Y.~Yan, J.~Cui, H.~Wang, Y.~Pang, W.~Jiang, J.~Zhang, Z.~Li, {\em et~al.}, ``Languagebind: Extending video-language pretraining to n-modality by language-based semantic alignment,'' {\em arXiv preprint arXiv:2310.01852}, 2023.

\end{thebibliography}
\end{document}